\documentclass{article}

\usepackage{microtype}
\usepackage{graphicx}
\usepackage{subfigure}
\usepackage{booktabs} 
\usepackage{pgfplots}
\pgfplotsset{compat=1.17}
\usepackage{hyperref}
\usepackage{xspace}

\def\genericmethodname{\textsc{DeltaLLM}\xspace}
\def\phicompressedname{\textsc{DeltaPhi}\xspace}
\def\llamacompressedname{\textsc{DeltaLlama}\xspace}

\usepackage[accepted]{icml2025}

\usepackage{amsmath}
\usepackage{amssymb}
\usepackage{mathtools}
\usepackage{amsthm}

\usepackage[capitalize,noabbrev]{cleveref}

\theoremstyle{plain}

\theoremstyle{definition}

\theoremstyle{remark}

\usepackage[textsize=tiny]{todonotes}

\icmltitlerunning{\genericmethodname: Compress LLMs with Low-Rank Deltas between Shared Weights}

\begin{document}

\twocolumn[
\icmltitle{\genericmethodname: Compress LLMs with Low-Rank Deltas between Shared Weights}



\icmlsetsymbol{equal}{*}

\begin{icmlauthorlist}
\icmlauthor{Liana Mikaelyan}{equal,yyy}
\icmlauthor{Ayyoob Imani}{equal,yyy}
\icmlauthor{Mathew Salvaris}{equal,yyy}
\icmlauthor{Parth Pathak}{equal,yyy}
\icmlauthor{Mohsen Fayyaz}{equal,yyy}
\end{icmlauthorlist}


\icmlaffiliation{yyy}{Microsoft}
\icmlcorrespondingauthor{Mohsen Fayyaz}{mohsenfayyaz@microsoft.com}

\icmlkeywords{Machine Learning, ICML}

\vskip 0.3in
]



\printAffiliationsAndNotice{\icmlEqualContribution} 

\begin{abstract} We introduce \genericmethodname, a new post-training compression technique to reduce the memory footprint of LLMs. We propose an alternative way of structuring LLMs with weight sharing between layers in subsequent Transformer blocks, along with additional low-rank difference matrices between them. For training, we adopt the progressing module replacement method and show that the lightweight training of the low-rank modules with approximately 30M-40M tokens is sufficient to achieve performance on par with LLMs of comparable sizes trained from scratch.

We release the resultant models, \llamacompressedname and \phicompressedname, with a 12\% parameter reduction, retaining 90\% of the performance of the base Llama and Phi models on common knowledge and reasoning benchmarks. Our method also outperforms compression techniques JointDrop, LaCo, ShortGPT and SliceGPT with the same number of parameters removed. For example, DeltaPhi 2.9B with a 24\% reduction achieves similar average zero-shot accuracies as recovery fine-tuned SlicedPhi 3.3B with a 12\% reduction, despite being approximately 400M parameters smaller with no fine-tuning applied.

This work provides new insights into LLM architecture design and compression methods when storage space is critical. \end{abstract}

\section{Introduction}

Transformer-based architectures~\cite{vaswani2017attention} have become the cornerstone of modern language modeling~\cite{abdin2024phi,dubey2024llama,team2024gemma,jiang2024mixtral,bai2023qwen,BrownGPT3}.  
While scaling laws indicate that performance improves as these models grow in size and training data~\cite{Kaplan20Scalinglaw,Hoffmann22Chinchila}, many model families are also trained in smaller sizes due to deployment constraints.  
In addition, multiple approaches---such as distillation~\cite{hinton2015distilling,gu2024minillm}, prompt and KV-cache compression~\cite{Zhang2023H2O,Cai2024PyramidKV,LiuDL2023Scissorhands,pan-etal-2024-llmlingua}, speculative decoding~\cite{LeviathanKM23specDecoding,LiW0024eagle2}, and quantization~\cite{ashkboos2024quarot,Frantar23gptq,xu2024onebit}---have been proposed to meet resource limitations.  
For extremely low-resource scenarios such as edge device deployments, these techniques are often combined to make deployment feasible.\footnote{\url{https://blogs.windows.com/windowsexperience/2024/12/06/phi-silica-small-but-mighty-on-device-slm/}}

Model compression has emerged as a prominent strategy to reduce both the model size and the computational overhead. 
Most previous works on compressing large language models (LLMs) rely on traditional pruning methods, which do not fully account for the multi-layered Transformer architecture at the core of these models~\cite{SinghAlistarh20,frantar2023sparsegpt,FrantarA22,Sun24pruning,Mishra2021,ma2023llmpruner}, resulting in suboptimal performance. 
Recently, several studies have proposed compression techniques more tightly tailored to LLMs~\cite{ashkboos2024slicegpt,men2024shortgpt,YangC024LaCo,he2024matterstransformersattentionneeded}.

\begin{figure}[t]
  \centering
  \includegraphics[width=0.4\textwidth]{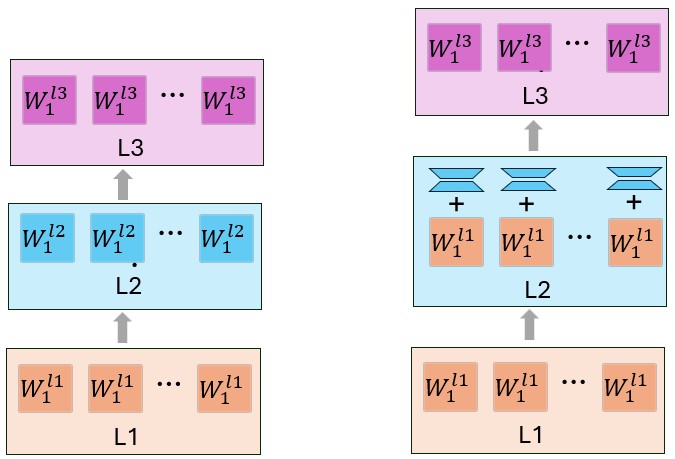} 
  \caption{\genericmethodname on the right, replaces some transformer layers with others, and account for the layer differences (delta) using low-rank matrices, which are trained to recover the original model's performance.}
   \label{fig:delta_figure}
\end{figure}

Although numerous post-training model compression methods have been studied independently, 
there is still no clear consensus on how to combine them for a specific deployment environment. 
We address this gap by systematically exploring several techniques 
including pruning~\cite{he2024matterstransformersattentionneeded}, 
knowledge distillation~\cite{hinton2015distilling}, 
weight sharing~\cite{dehghani2018universal}, 
low-rank adaptation~\cite{hu2021lora}, 
and progressive module replacement~\cite{xu2020berttheseus}. 
Based on the insights gained, we introduce \genericmethodname{}, a unified compression strategy 
that can create significantly smaller language models. 
Our experiments show that using only 37M tokens for the training of the compressed model, 
\genericmethodname{} outperforms training a new model from scratch even with billions of tokens with similar size, 
while preserving competitive performance.

\genericmethodname{} draws inspiration from prior work suggesting redundancy between Transformer layers~\citep{he2024matterstransformersattentionneeded, YangC024LaCo, men2024shortgpt}, 
as well as from studies showing that simply repeating layers (weight sharing) is an effective method to increase model depth~\citep{dehghani2018universal, MobileLLM, Lan2020ALBERT}. 
In our approach, we replace a Transformer layer with another one, but introduce low-rank ``delta weights'' to account for the minor differences between shared layers (see Figure~\ref{fig:delta_figure}). 
Rather than storing new parameters for every Transformer block, we thus store only a smaller subset of unique weights, complemented by learnable low-rank updates, 
significantly reducing overall model size.

We train low-rank matrices to recover the original model's performance, using knowledge distillation~\cite{hinton2015distilling} with the original LLM as the teacher. To ensure smooth layer substitution, we follow a progressive module replacement strategy, previously shown effective for compressing encoder-only language models~\cite{xu2020berttheseus}. We conduct ablation studies on both our architectural and training choices, demonstrating their effectiveness. Additionally, our approach is orthogonal to pruning and quantization, yielding hardware-friendly compressed models that achieve performance comparable to the original models.




Our contributions are as follows:
\begin{itemize}
    \item We present \genericmethodname, a new way of structuring Transformer-based models that focuses on weight sharing and low-rank differences between Transformer blocks, significantly reducing the memory requirements for on-device use cases.
     \item We use \genericmethodname to compress the Phi-3.5 and Llama-3.2 families of models and achieve better performance on common reasoning and knowledge benchmarks than existing SLMs of comparable sizes. We further show that \phicompressedname and \llamacompressedname outperform post-training compression methods JointDrop, SliceGPT, ShortGPT, and LaCO with the same number of parameters removed. 
    \item To the best of our knowledge, we are the first to apply a progressive module replacement method to distill decoder-only LLMs. We show that it converges faster than standard knowledge distillation. We further propose uneven replacement probabilities across layers. 
    

    
\end{itemize}

We hope that our observations provide valuable insights on Transformer-based model redundancies, distillation strategies and best practices for LLM architecture design, specifically when storage space is critical.


\section{Related Work}

\textbf{Transformers Redundancy}

The lottery ticket hypothesis posits that a dense, randomly initialized model contains subnetworks (or ``winning tickets'') which, when trained in isolation, can match or exceed the accuracy of the original model~\citep{frankle2018the}. This hypothesis has also been validated in reinforcement learning and natural language processing~\citep{Yu2020Playing, DBLP:conf/icml/FrankleD0C20}. Recent studies further reveal substantial redundancy in Transformer architectures~\citep{men2024shortgpt}, with efforts focusing on removing entire blocks or selectively dropping attention and MLP layers using similarity metrics~\citep{he2024matterstransformersattentionneeded, deeperLook, bian-etal-2021-attention}. Additionally, there is evidence that redundancy tends to concentrate in the middle to later layers, while the first and last layers play more specialized roles~\citep{men2024shortgpt, ma2023llmpruner}.

\textbf{Pruning}


Pruning has emerged as a widely used model compression technique, removing less important parameters to reduce computational and memory footprints~\cite{Han2015}. Early efforts often relied on unstructured pruning~\citep{han2015deep,h.2018to,Trevor2019sparsity,FrantarA22,FrantarSparseGPT}, which did not necessarily improve inference speed. Consequently, more recent approaches explored structured pruning—removing entire filters or attention heads for predictable speedups~\cite{HoeflerABDP21,Mishra2021,Sun24pruning}. 

In large Transformer-based models, specialized strategies typically apply structured pruning followed by continued training. For example, movement pruning~\cite{Sanh0R20} dynamically drops redundant connections during fine-tuning; \citet{ashkboos2024slicegpt} uses principal component analysis to remove trailing rows and columns of weight matrices; \citet{ma2023llmpruner} selects non-critical coupled structures based on gradient information; \citet{men2024shortgpt} eliminates whole layers based on input--output similarity; and LaCo~\cite{YangC024LaCo} collapses outermost layers into their neighbors. MINITRON~\cite{muralidharan2024compact} further compares various pruning and re-training approaches, demonstrating superior efficiency over training small language models from scratch. An essential difference between MINITRON and \genericmethodname{} is the training data requirement: while \genericmethodname{} recovers performance with as few as 37M tokens, MINITRON requires billions.

\textbf{Weight Sharing}

Weight sharing, which offers flexible ways to reduce computational and memory overhead, has recently attracted significant attention. Two widely adopted weight-sharing approaches are Mixture of Experts (MoE) and cross-layer weight sharing. MoE, originally introduced to machine learning as a method similar to ensemble techniques~\cite{JacobsJNH91MOE}, has since been applied to various architectures~\cite{fedus2022reviewMoE,EigenRS13MoE,ShazeerMMDLHD17MoE} including Transformers~\cite{jiang2024mixtral,dubey2024llama,abdin2024phi}. Cross-layer weight sharing, introduced for Transformer-based language modeling in~\cite{dehghani2018universal} as a way to reduce memory usage and improve performance, was subsequently adopted by multiple works for training Transformer models from scratch~\cite{wang2024residualtransformer,dehghani2018universal,Lan2020ALBERT,MobileLLM} or for shrinking existing models~\cite{bae2024relaxed}.

\textbf{Low-Rank Adaptation}

Low-rank approximations (LoRA) have been extensively applied to various stages of machine learning for diverse purposes. Aside from tasks that inherently rely on low-rank structures \citep{cai2010singular,li2016recovery,li2018algorithmic}, LoRA-based methods have been proposed to improve the generalization of over-parameterized models \citep{oymak2019generalization}, enable efficient model fine-tuning \citep{hu2022lora,hyeon2021fedpara,yeh2024navigating,Mahabadi21compacter}, and reduce computational costs by integrating low-rank layers directly into model architectures \citep{Lan2020ALBERT} or as a post-training compression step \citep{ben-noach-goldberg-2020-compressing,tukan2020compressed,NIPS2014_2afe4567}. 

\begin{figure}
    \centering
    \includegraphics[width=1\linewidth]{    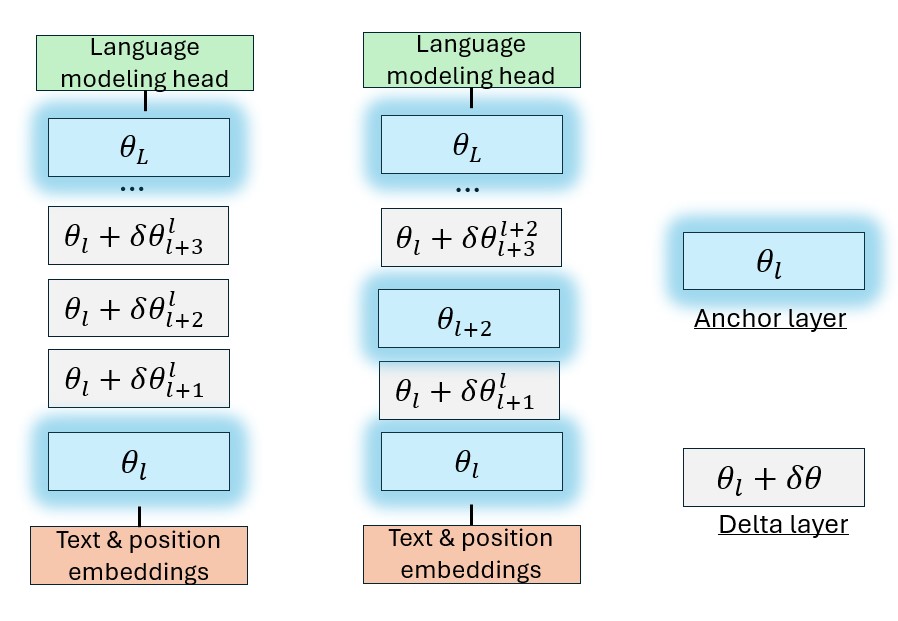}
    \caption{Two ways to structure for a Delta-Model: delta-layers at each subsequent block after a base block (left) and alternating blocks with delta modules between (right).}
    \label{fig:deltaLLM}
\end{figure}







\section{Methodology}

\subsection{Preliminaries}

For a weight $\mathbf{W}^{D\times D}$ the LoRA technique introduces two additional low-rank matrices $\mathbf{A}^{D \times R}$ and $\mathbf{B}^{R \times D}$, where $R$ is the rank of the matrices. These matrices are then updated during the fine-tuning process and are added to the weight, while the rest of the network remains frozen.

\subsection{\genericmethodname}
\label{sec:method_detail}
\genericmethodname 
introduces low-rank differences between layers in consecutive Transformer blocks that share weights, called \textit{\textbf{deltas}}. 

A model weight $\mathbf{W}$ of layer $l+i$ can be restructured as a function of the previous layer $l$ and a \textit{\textbf{delta}} between the two weights as follows:
\begin{equation}\label{eq:delta1}
\mathbf{W}_{l+i}^{M\times N} = \mathbf{W}_l^{M\times N} + \tilde{\delta}_{l+i}^{l}
\end{equation}
where $\tilde{\delta}_{l+i}^{l}$ is the low-rank approximation of
\begin{equation}\label{eq:delta2}
\delta_{l+i}^{l} = \mathbf{W}_{l+i}^{M\times N} - \mathbf{W}_l^{M\times N}
\end{equation}
We define the delta matrices in the same manner as the $A$ and $B$ matrices in the LoRA setup.
The low-rank approximations can be obtained using Singular Value Decomposition (SVD) or other approximation methods. The low-rank deltas allow to restore diversity and adaptiveness that get affected due to layer replication. 

Refer to figure \ref{fig:deltaLLM} for the architecture of a \genericmethodname model. We present two strategies for organizing the model structure: a single Transformer block with delta modules creating each subsequent block (left) and alternating Transformer blocks with delta modules in between (right). Within blocks, we can compress attention and/or multi-layer perceptron (MLP) layers. 

We can further extend this to allow any previous layer's weight $\mathbf{W_{l-k}}$ to be used to initialize the current weight $\mathbf{W_l}$. We refer to these layers as \textbf{\textit{anchor}} layer and the corresponding Transformer blocks as \textbf{\textit{anchor}} blocks.

Refer to the ablation studies in Section 5.7 for the best practices on the choice of the blocks and layers. 

\subsection{Delta-Model Training}
While with the right initialization the deltas may be sufficient to retain the desired performance, our method relies on further training, albeit on a small number of tokens. 

We explore two strategies for training a \genericmethodname model:
\begin{itemize}
    \item \textbf{Delta-tuning only with Progressive Module Replacement}: delta-layers are progressively replaced with original layers. Only the delta-layers are trained while the rest of the model weights remain fixed.
    \item \textbf{Delta-layer tuning with LoRA fine-tuning}: LLM weights are fine-tuned jointly with the delta weights using parameter-efficient training methods.
\end{itemize}

\textbf{Delta-tuning only}


Following \cite{xu2020berttheseus}, we progressively replace original LLM weights with the corresponding \genericmethodname weights according to a probability scheduler. That is, at the beginning of training, modules are replaced with the given probability rate which gradually increases until convergence to 1.0. Unlike \cite{xu2020berttheseus}, after probability rate convergence, we continue training for an additional number of epochs, as determined by the hyperparameter search we conduct. 


We additionally extend the progressive module replacement (PMR) method to allow uneven replacement rates across the blocks. This is motivated by our preliminary experiments that later layers contain more redundancy than earlier layers and that it may be beneficial to replace the later layers first for smoother training. 


The total loss is computed as 
\begin{equation}\label{eq:delta1}
L = (1-\alpha ) L_{CE} + \alpha L_{logits},
\end{equation}
where $L_{CE}$ is the cross entropy of the student model (with potentially some layers replaced with the teacher model), $L_{logits}$ is the distillation loss between the final logits of the teacher and the student models. Following \cite{muralidharan2024compact}, we choose Kullback–Leibler (KL) divergence as the distillation loss as it is shown to outperform the mean squared error (MSE) and the cosine similarity, $\alpha$ is the distillation weight.

\textbf{Full model Fine-tuning}

In the second approach for \genericmethodname model training, we train the delta-layers using the progressing module replacement along with the fine-tuning of the rest of the model. We adopt parameter-efficient fine-tuning methods.

\section{Experiments}

\subsection{Experiment Settings}

\begin{table*}[t]
\caption{Zero-shot benchmark evaluation results of Delta-Models and pre-trained models of similar sizes.}
\label{lm_eval}
\vskip 0.15in
\begin{center}
\begin{small}
\begin{tabular}{lccc|ccccr}
\toprule
\textbf{Benchmark} & \multicolumn{5}{c}{\textbf{Models}} \\
\midrule
 &  \textbf{Phi 3.5} & \textbf{Llama 3.2} &  \textbf{Qwen 2.5} &\textbf{\phicompressedname} & \textbf{\phicompressedname} & \textbf{\llamacompressedname} & \textbf{\llamacompressedname}\\
\midrule
\# Parameters & 3.8B & 3.2B & 3.2B & 3.35B & 2.9B & 2.52B & 2.41B & \\
Compression \% & & & & 12\% & 24\% & 21\% & 25\% \\
Base Model & & & & Phi 3.5 & Phi 3.5 & Llama 3.2 & Llama 3.2 \\
Training Stage & & & & $\delta$-tuned only & $\delta$-tuned only & $\delta$-tuned only &$\delta$-tuned only\\
Training tokens & & & & 37M & 37M & 32M & 32M \\
Delta-layers rank/size & & & & r1000/580MB & r100/90MB & r1000/380MB & r100/30MB \\
\midrule
MMLU\_pro (acc) & \textbf{0.36} & 0.29 & 0.32  & \textbf{0.32} & 0.31 & 0.28 & 0.26 \\
WG (acc) & \textbf{0.75} & 0.67 & 0.69 & 0.70 & \textbf{0.71} & 0.65 & 0.61 \\
ARC\_c (acc\_norm) & \textbf{0.61} & 0.46 & 0.47 & \textbf{0.51} & 0.44 & 0.35& 0.34 \\
HS (acc\_norm) & \textbf{0.77} & 0.70 & 0.74 & \textbf{0.70} & 0.61 & 0.59 & 0.55 \\
PIQA (acc\_norm) & \textbf{0.80} & 0.76 & 0.74 & \textbf{0.74} & 0.72 & 0.70 & 0.69 \\ 
\midrule
Average & \textbf{0.66} & 0.58 & 0.59 & 0.59 & 0.56 & 0.51 & 0.50 \\
\bottomrule
\end{tabular}
\end{small}
\end{center}
\vskip -0.1in
\end{table*}

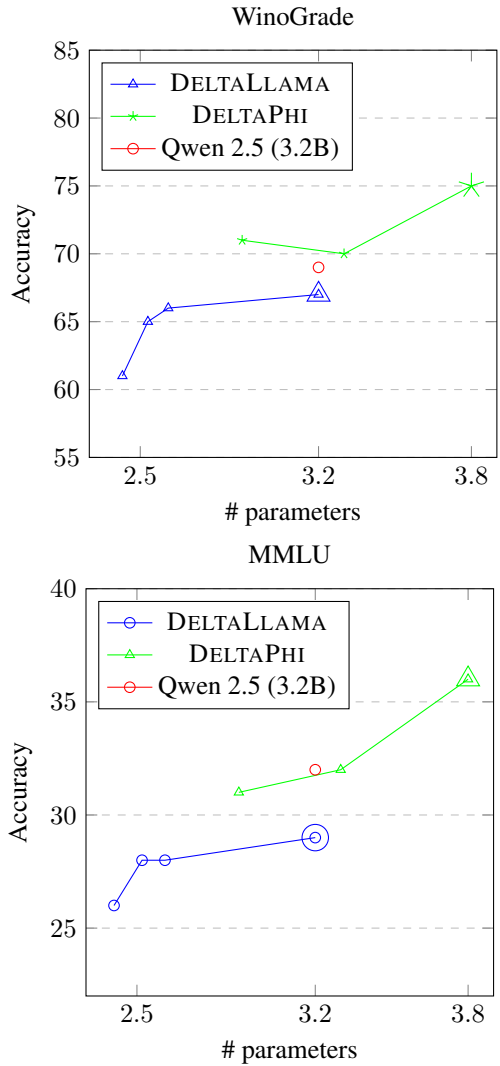
\begin{figure}[t]
    \centering
    \begin{tikzpicture}
        \begin{axis}[
            title={WinoGrade},
            width=7cm,
            height=7cm,
            xlabel={\# parameters},
            ylabel={Accuracy},
            xmin=2.3, xmax=3.9,
            ymin=55, ymax=85,
            xtick={2.5, 3.2, 3.8},
            ytick={50,55,60,65, 70,75, 80,85},
            legend pos=north west,
            ymajorgrids=true,
            grid style=dashed,
        ]
        \addplot[
            color=blue,
            mark=triangle,
            ]
            coordinates {
            (3.2,67)(2.61,66)(2.53,65)(2.43,61)
            };
        \addplot[
            color=green,
            mark=star,
            ]
            coordinates {
            (3.8,75)(3.3,70)(2.9,71)
            };
        \addplot[
            color=red,
            mark=o,
            ]
            coordinates {
            (3.2,69)
            };
        \addplot[
            color=green,
            mark=star,
            mark size=5pt,
            ]
            coordinates {
            (3.8,75)
            };
         \addplot[
            color=blue,
            mark=triangle,
            mark size=5pt,
            ]
            coordinates {
            (3.2,67)
            };
            
        \legend{\llamacompressedname, \phicompressedname, Qwen 2.5 (3.2B)}
        \end{axis}
    \end{tikzpicture}
    \hspace{1cm}
    \begin{tikzpicture}
        \begin{axis}[
            title={MMLU},
            width=7cm,
            height=7cm,
            xlabel={\# parameters},
            ylabel={Accuracy},
            xmin=2.3, xmax=3.9,
            ymin=22, ymax=40,
            xtick={2.5, 3.2, 3.8},
            ytick={20,25, 30, 35, 40},
            legend pos=north west,
            ymajorgrids=true,
            grid style=dashed,
        ]
        \addplot[
            color=blue,
            mark=o,
            ]
            coordinates {
            (3.2,29)(2.61,28)(2.52,28)(2.41,26)
            };
        \addplot[
            color=green,
            mark=triangle,
            ]
            coordinates {
            (3.8,36)(3.3,32)(2.9,31)
            };
        \addplot[
            color=red,
            mark=o,
            ]
            coordinates {
            (3.2,32)
            };
        \addplot[
            color=green,
            mark=triangle,
            mark size=5pt,
            ]
            coordinates {
            (3.8,36)
            };
        \addplot[
            color=blue,
            mark=o,
            mark size=5pt,
            ]
            coordinates {
                (3.2,29)
                };
                \legend{\llamacompressedname, \phicompressedname, Qwen 2.5 (3.2B)}
                \end{axis}
                \end{tikzpicture}
                \caption{DeltaLlama and DeltaPhi accuracies on WinoGrade and MMLU pro}
    \label{fig:MMLU and WinoGrade}
\end{figure}

We use Phi 3.5 \cite{abdin2024phi} and Llama 3.2 \cite{dubey2024llama} as the teacher models to obtain \phicompressedname and \llamacompressedname models respectively using the procedure outlined in Section 4.

We use Alpaca \cite{alpaca} and Ultrachat \cite{ding2023enhancing} for all training experiments with \phicompressedname and \llamacompressedname. Alpaca is separated into train (80\%), validation (10\%) and test (10\%) portions. The train set is used to conduct the hyperparameter search, for which the test set is used as an optimization metric. The Delta-Model is then trained on the Ultrachat train set with the best set of the hyperparameters. 

We explore four types of initializing the delta-layers:
\begin{itemize}
    \item \textbf{Gaussian}: initializes the A matrix with a Gaussian distribution and the B matrix as zero.
    \item \textbf{PiSS}A: performs SVD on the weight matrix and chooses the principal singular values and vectors to initialize both A and B \cite{meng2024pissa}.
    \item \textbf{OLoRA}: performs QR decomposition to initialize A and B \cite{buyukakyuz2024olora}. 
    \item \textbf{EVA}: a data-driven initialization method that performs SVD on input activations of every layer and uses the right-singular vectors to initialize the LoRA matrices \cite{paischer2024one}.
\end{itemize}

We adopt the PEFT library for initializing the delta layers \cite{peft}. 
All our experiments are conducted on A100 80GB Nvidia GPUs.

We observe that the training stability is very sensitive to the hyperparameters, thus, an extensive search is conducted using the perplexity on Alpaca test set as the optimization metric. We adopt the Bayesian Optimization search strategy provided by the SyneTune library \cite{salinas2022syne}. We search over the learning rate, learning rate scheduler, number of epochs until convergence, the number of extra epochs, distillation weight and type, module replacement probability, delta-layers initialization, LoRA $\alpha$ and LoRA dropout. The search is conducted with a maximum number of jobs of 40.

\subsection{Commonsense benchmark evaluation}

We evaluate \phicompressedname and \llamacompressedname models on common knowledge and reasoning benchmarks including MMLU-Pro \cite{wang2024mmlu}, WinoGrande \cite{sakaguchi2021winogrande}, HellaSwag \cite{zellers-etal-2019-hellaswag}, ARC-Challenge \cite{clark2018think}. We adopt Language Model Evaluation Harness framework \cite{eval-harness} to perform the evaluations.
We conduct zero-shot evaluation for all benchmarks. 

The strategy 1 experiment results are shown in Table \ref{lm_eval} and Table \ref{Stage1}. Here we compare several \phicompressedname and \llamacompressedname models with the teacher models they were derived from as well as models of similar sizes. Observe that \phicompressedname 3.35B outperforms Llama 3.2B and Qwen 3.2B on MMLU, WinoGrande and ARC-Challenge benchmarks with a comparable average accuracy across all the tasks. In addition, \phicompressedname 3.35B and \phicompressedname 2.9B achieve similar results for MMLU and WinoGrande results despite the later being compressed by 24\% as opposed to 12\% and being $\sim 400$M parameters smaller.

Note that for generation tasks with perplexity as the metric (Table \ref{Stage1} in the Appendix) on large datasets (Ultrachat), \phicompressedname outperforms the original Phi 3.5 it was compressed from. In addition, \phicompressedname achieves lower perplexity than Llama 3.2 models of comparable sizes.

Also note that the delta-layers in the \phicompressedname 2.9B model account for only 90MB which is significantly lower than the storage reduced by layer replication.







\begin{table*}[t]
\caption{Comparison of \phicompressedname with SOTA compression methods}
\label{baseline_phi}
\vskip 0.15in
\begin{center}
\begin{small}
\begin{tabular}{lccccccr}
\toprule
Benchmark & \multicolumn{6}{c}{Models} \\
\midrule
  & \textbf{\phicompressedname} & \textbf{\phicompressedname}  & \textbf{DropPhi} & \textbf{SlicedPhi} & \textbf{ShortPhi} & \textbf{Phi-LaCo} \\
\midrule
\# Parameters & 3.35B & 2.9B & 3.4B & 3.32B & 3.37B & 3.36B \\
Method & \genericmethodname & \genericmethodname & JointDrop & SliceGPT  & ShortGPT & LaCo \\
Compression & 12\% & 24\% & 11\% & 12\% & 12\% & 12\% \\
Train Tokens & 37M & 37M & & 37M &  \\
Train Stage & $\delta$-tuned only & $\delta$-tuned only  & No training & LoRA fine-tuned & No training & No training \\
\midrule
MMLU\_pro (acc)  & 0.32  &  0.31 & 0.28  & 0.24 & 0.21 & \textbf{0.33} \\
WG (acc)   & 0.70 & \textbf{0.71} & 0.60  & 0.68 & 0.64 & \textbf{0.71}\\
ARC\_c (acc\_norm)  & \textbf{0.51} & 0.44 & 0.44 & 0.49 &  0.46 & 0.50 \\
 HS (acc\_norm) & \textbf{0.70} & 0.61 & 0.53 & 0.65 & 0.68 & 0.63 \\
PIQA (acc\_norm) & 0.74 & 0.72 & 0.67 & 0.74 & \textbf{0.76} & 0.73  \\
\midrule
Average & \textbf{0.59} & 0.56 & 0.50 & 0.56 & 0.55 & 0.58 \\
\bottomrule
\end{tabular}
\end{small}
\end{center}
\vskip -0.1in
\end{table*}

\begin{table*}[t]
\caption{Comparison of \llamacompressedname with SOTA compression methods}
\label{baseline_llama}
\vskip 0.15in
\begin{center}
\begin{small}
\begin{tabular}{lcccccr}
\toprule
Benchmark & \multicolumn{6}{c}{Models} \\
\midrule
& \textbf{\llamacompressedname} &  \textbf{\llamacompressedname} &\textbf{DropLlama} & \textbf{SlicedLlama} & \textbf{ShortLlama} & \textbf{Llama-LaCo} \\
\midrule
\# Parameters & 2.52B & 2.41B & 2.6B & 2.56B & 2.61B & 2.41B \\
Method & \genericmethodname & \genericmethodname & JointDrop & SliceGPT & ShortGPT & LaCo \\
Compression & 21\% & 25\% & 19\% & 20\%& 18\% & 25\% \\
Train Tokens & 32M & 32M & & 32M & \\
Train Stage & $\delta$-tuned only & $\delta$-tuned only & No training & LoRA fine-tuned & No training & No training \\
\midrule
MMLU\_pro (acc)  & \textbf{0.28} & 0.26 & 0.11 & 0.11 & 0.23 & 0.15  \\
WG (acc)    & \textbf{0.65} & 0.61 & 0.64 & 0.49 & 0.64 & 0.63 \\
ARC\_c (acc\_norm)  & 0.35 & 0.34 & \textbf{0.40} & 0.26 & 0.36 & 0.32 \\
HS (acc\_norm)  & \textbf{0.59} & 0.55 &  0.58 & 0.28 & 0.56 & 0.52 \\
PIQA (acc\_norm) & \textbf{0.70} & 0.69 & 0.65 & 0.55 & 0.64& 0.65 \\
\midrule
Average & \textbf{0.51} & 0.50 & 0.48 & 0.33 & 0.49 & 0.45 \\
\bottomrule
\end{tabular}
\end{small}
\end{center}
\vskip -0.1in
\end{table*}

\subsection{Model quantization}
 We conduct 8-bit and 4-bit quantization experiments on a \phicompressedname model with 2.9 billion parameters. Specifically, we employ llm.int8 \cite{dettmers2022gpt3} method, which applies weight quantization and dynamically determines the activation scales to minimize the information loss. Only the base model undergoes quantization, while the delta layers remain in FP16 to preserve adaptation capacity. 
 
 The model architecture incorporates weight sharing in its MLP layers, specifically between blocks 22–29 and block 21, with these shared layers connected via rank-100 delta matrices.

 For quantization, we use an 8-bit scheme based on vector-wise quantization of weights, which balances compression and inference efficiency. For 4-bit quantization, we use the NF4 data format, which is a non-uniform 4-bit quantization scheme. Both methods leverage mixed-precision decomposition techniques to maintain numerical stability.

Our goal is to evaluate whether quantization has any adverse effect on model performance and, subsequently, whether excluding the anchor layers from quantization impacts the overall quantization efficiency. Hence, we explore two distinct quantization strategies for this architecture:

\begin{itemize} \item \textbf{AnchorSkip}: Quantize all layers except the anchor layers. \item \textbf{AllQuant}: Quantize all layers, including the anchor layers. \end{itemize}

The results in Table~\ref{quant} show that quantization only minimally degrades the model performance when compared with the non-quantized model. There is also marginal gain using AnchorSkip strategy compared to the AllQuant.

\begin{table}[t]
\caption{Quantization Results on \phicompressedname 2.9B}
\label{quant}
\begin{center}
\begin{small}
\begin{tabular}{lcc}
\toprule
\textbf{Dataset} & \textbf{AllQuant} & \textbf{AnchorSkip} \\
\midrule
\multicolumn{3}{l}{\textbf{8-bit Quantization}} \\
\hspace{1em} MMLU\_pro (acc)   &     0.30      &    0.33        \\

\hspace{1em} HS (acc\_norm)     &     0.60      &    0.61       \\
\hspace{1em} ARC\_c (acc\_norm) &     0.42      &    0.44        \\
\hspace{1em} PIQA (acc\_norm)   &     0.72      &    0.72        \\
\midrule
\multicolumn{3}{l}{\textbf{4-bit Quantization}} \\
\hspace{1em} MMLU\_pro (acc)   &     0.28      &   0.29        \\
\hspace{1em} HS (acc\_norm)     &     0.59      &   0.60         \\
\hspace{1em} ARC\_c (acc\_norm)    &  0.42      &   0.42               \\
\hspace{1em} PIQA (acc\_norm)   &     0.72      &   0.72         \\
\bottomrule
\end{tabular}
\end{small}
\end{center}
\vspace{-0.3cm} 
\end{table}

\subsection{Ablation Study: Progressive Module Replacement}

In this section we compare $\delta$-only training with PMR and the standard distillation where the teacher and student models are completely separate. This is equivalent to setting the replacement rate to 1.0 while keeping the distillation loss unchanged.
Refer to Table \ref{ablation1} for the results on Alpaca test dataset. With extensive hyperparameter optimization including search over the number of epochs, the PMR approach outperforms the distillation without PMR with a smaller number of epochs. This suggests that PMR enables faster convergence.

\begin{table}[t]
\caption{Comparison of distillation with PMR and constant replacement rate of 1.}
\label{ablation1}
\begin{center}
\begin{small}
\begin{tabular}{lccccr}
\toprule
\textbf{Model} & \textbf{Distil Method} & \textbf{Epochs} & \textbf{Dataset} & \textbf{PPL} \\
\midrule
\phicompressedname 3B & PMR & \textbf{2} & Alpaca & \textbf{3.34} \\
\phicompressedname 3B & without PMR & 5 & Alpaca & 3.42 \\
\phicompressedname 3B & with PMR & \textbf{2} & Ultrachat &  7.49 \\
\phicompressedname 3B & without PMR & 4 & Ultrachat & \textbf{7.24} \\
\bottomrule
\end{tabular}
\end{small}
\end{center}
\end{table}

\subsection{Ablation Study: Which blocks and layers to compress?}

In all our experiments we keep the first and the last two blocks unchanged since they are highly specialized and cause significant degradation if removed \citep{men2024shortgpt, ma2023llmpruner}.

We explore three strategies of the construction of the delta-blocks:

\begin{itemize}
    \item \textbf{Sequential}: A sequential set of delta-blocks. That is, weights of a single anchor Transformer block get shared with the next few blocks with deltas.
    \item \textbf{Alternating}: A set of blocks where a standard and delta block are alternating. In this setup multiple anchor blocks share weight with other blocks and have deltas between them.
    \item \textbf{JointDrop}: The layers obtained by the JointDrop method \cite{he2024matterstransformersattentionneeded}. The JointDrop method computes the importance of layers using a similarity-based metric to capture redundancy. This metric is then used to eliminate those layers completely. In our setup, instead of removing them, we replicated those layers with the addition of deltas.
\end{itemize}

In each of these options, we can further choose the layers to compress within the blocks: MLP, attention or a combination of both.
When choosing which layers to replicate the weights from, we rely on
prior work suggesting that the changes between the outputs of consecutive layers tends to be less significant \cite{YangC024LaCo}. Hence, we always choose the anchor layer to be layer prior to the delta-layer.


Refer to Table \ref{ablation2} for the ablations on the Transformer blocks chosen for Phi-3.5. Note that the JointDrop method always outputs the attention layers. This happens due to the importance computation of the inputs and outputs of the layers.

In the sequential and alternating cases, the MLP choice tends to results in lower perplexities for the models with more layers removed.

We hypothesize that the compression of MLP layers tends to perform better due to the nature of the \genericmethodname architecture. Attention layers are highly sensitive to the input tokens and their replication may lead to the loss of layer-specific toke-to-token relationships, hurting effective language modeling.
MLP layers transform tokens independently and contribute to most of the model parameters, suggesting that they may exhibit more redundancy. Our method does not remove the MLP layers as pruning methods do, preserving the number of computations performed and the trained low-rank deltas allow to diversify the replicated MLPs.

In addition, given the same number of blocks, compressing MLP layers helps to reduce the storage requirements more significantly due to the large number of parameters.
Hence, in the subsequent experiments we apply our method to the MLP layers.  

\begin{table}
\caption{\phicompressedname ablations with block/layer choice}
\label{ablation2}
\begin{center}
\begin{small}
\begin{tabular}{lcccr}
\toprule
\textbf{Model} & \textbf{Block Choice} &  \textbf{Dataset} & \textbf{PPL} \\
\midrule
\phicompressedname (4 ATTN)  & JointDrop & Alpaca & \textbf{3.14} \\
\phicompressedname (6 ATTN)  & JointDrop & Alpaca & 3.76 \\
\phicompressedname (8 ATTN)  & JointDrop & Alpaca & 4.27 \\
\midrule
\phicompressedname (4 MLP)  & Sequential & Alpaca & 3.24 \\
\phicompressedname (6 MLP)  & Sequential &  Alpaca & 3.80 \\
\phicompressedname (8 MLP)  & Sequential &  Alpaca & \textbf{3.94} \\
\midrule
\phicompressedname (4 ATTN)  & Sequential & Alpaca & 3.29 \\
\phicompressedname (6 ATTN)  & Sequential &  Alpaca & 3.84 \\
\phicompressedname (8 ATTN)  & Sequential &  Alpaca & 4.31 \\
\midrule
\phicompressedname (4 MLP)  & Alternating &  Alpaca & 3.22 \\
\phicompressedname (6 MLP)  & Alternating &  Alpaca & \textbf{3.55} \\
\phicompressedname (8 MLP)  & Alternating &  Alpaca & 4.10 \\
\midrule
\phicompressedname (4 ATTN)  & Alternating &  Alpaca &  3.36 \\
\phicompressedname (6 ATTN)  & Alternating &  Alpaca &  3.58 \\
\phicompressedname (8 ATTN)  & Alternating &  Alpaca &  4.03 \\
\bottomrule
\end{tabular}
\end{small}
\end{center}
\end{table}







\subsection{Baselines} 

In this section, we compare our method to state-of-the-art compression methods JointDrop, SliceGPT, ShortGPT and LaCo. The results are given in the Tables \ref{baseline_phi} and \ref{baseline_llama}.

\textbf{JointDrop: Removing Attention and MLP layers}

\cite{he2024matterstransformersattentionneeded} show that several layers in an LLM can be completely removed while maintaining comparable model performance. We replicate their method on Phi 3.5 and Llama 3.2.

\genericmethodname with comparable number of parameters outperforms Drop-Phi on many of the common benchmarks.

Interestingly, we observe that smaller models like Phi 3.5 experience more significant performance drops during layer removal compared to the larger models (LLama 70B) examined in the original study. This aligns with other works suggesting that larger models contain more redundancy than smaller models. 
This may also suggest that model compression techniques may need to be specifically tailored to the scale of the target model rather than applying a one-size-fits-all approach.
Due to the resource constraints we did not reproduce the results on larger models.

\textbf{Pruning Methods}


We compare our approach to the SliceGPT pruning method with similar number of parameter removed. Note that for SliceGPT experiments, the chosen sparsity level does not correspond to the same percentage reduction in the number of model parameters: e.g. we produced SlicedPhi with 23\% Transofrmer-block sparsity resulting in a model compressed by 13\% (total sparsity). For \genericmethodname models, the reported compression percentage accounts for both the removed layers and the additional delta matrices.
We further apply recovery fine-tuning on the sliced models using the default parameters. 

\phicompressedname and \llamacompressedname with delta-tuned weights consistently outperform SlicedPhi and SlicedLlama of comparable sizes respectively on the common benchmarks, despite no recovery fine-tuning of the \genericmethodname models. 

Tables \ref{baseline_phi} and \ref{baseline_llama} also present results with the SOTA pruning method LaCo. Note that LaCo on Phi tends to perform well, outperforming \phicompressedname on MMLU and Winogrande, however the average accuracy across the benchmarks is on par with \phicompressedname. LaCo applied to Llama produces significantly worse results than \llamacompressedname. 
Overall, \phicompressedname and \llamacompressedname have the highest average accuracy compared to models obtained using other post-training compression methods. We hypothesize that the trained deltas are able to capture the slight differences between the consecutive shared weights. All of the other methods reduce the overall number of computations unlike our method.



\section{Limitations and Future Work}

\textbf{Delta-layer Initialization}

Compared with training-free compression methods, our method heavily relies on the training of the low-rank deltas. This implies a dependency on the quality of the training data as well as the need for additional compute resources, although light. We hypothesize that the right initialization of deltas may reduce the need for additional training. We plan to explore this as a future research direction.

\textbf{Shared Layer Operations}

In the current setup, the subsequent layers are initialized with full weight replication of the anchor weights. We hypothesize that the shared weights can also be initialized using other operations such as a weighted average of the weight matrices. This may facilitate a better knowledge transfer between the layers.



\textbf{Inference latency}

The \genericmethodname currently focuses on restructuring an LLM to save disk space on-device while preserving comparable accuracy. Our method does not focus on reducing the inference time of the model running on-device. We plan to investigate ways to reduce the inference time in future work. For example, a potential approach is to combine attention pruning or removal methods with MLP weight sharing, which can contribute to both space and computation efficiency. 





\section{Conclusion}
In this paper, we present \genericmethodname, an alternative approach to structuring Transformer-based models, which optimize for space efficiency. In our setting, MLP and attention layers share weights with the corresponding anchor layers earlier in the network. The shared weights have additional low-rank delta-layers between them, which are trained to preserve the knowledge and capabilities of the model. We show that \genericmethodname models compressed from open-source LLMs can achieve comparable performance with other models of similar sizes trained from scratch.
This new structure also outperforms many post-training compression methods such as JointDrop, SliceGPT, ShortGPT and LaCo. We hope that our method inspires new efficient designs of LLM architectures that can be trained from scratch as well as serve as both a post-training compression technique. 

\section{Impact Statements}

This paper presents work whose goal is to advance the field of Machine Learning. There are many potential societal consequences of our work, none which we feel must be specifically highlighted here.

\section*{Acknowledgements}
We would like to thank James Hensman, Taketomo Isazawa, Mohamed Nour Abouelseoud, Pashmina Cameron, Sunando Sengupta and Eric Sommerlade for their helpful feedback on the paper.

\bibliography{icml_2025}

\begin{thebibliography}{78}
\providecommand{\natexlab}[1]{#1}
\providecommand{\url}[1]{\texttt{#1}}
\expandafter\ifx\csname urlstyle\endcsname\relax
  \providecommand{\doi}[1]{doi: #1}\else
  \providecommand{\doi}{doi: \begingroup \urlstyle{rm}\Url}\fi

\bibitem[Abdin et~al.(2024)Abdin, Aneja, Awadalla, Awadallah, Awan, Bach, Bahree, Bakhtiari, Bao, Behl, et~al.]{abdin2024phi}
Abdin, M., Aneja, J., Awadalla, H., Awadallah, A., Awan, A.~A., Bach, N., Bahree, A., Bakhtiari, A., Bao, J., Behl, H., et~al.
\newblock Phi-3 technical report: A highly capable language model locally on your phone.
\newblock \emph{arXiv preprint arXiv:2404.14219}, 2024.

\bibitem[Ashkboos et~al.(2024{\natexlab{a}})Ashkboos, Croci, Nascimento, Hoefler, and Hensman]{ashkboos2024slicegpt}
Ashkboos, S., Croci, M.~L., Nascimento, M. G.~d., Hoefler, T., and Hensman, J.
\newblock Slicegpt: Compress large language models by deleting rows and columns.
\newblock \emph{arXiv preprint arXiv:2401.15024}, 2024{\natexlab{a}}.

\bibitem[Ashkboos et~al.(2024{\natexlab{b}})Ashkboos, Mohtashami, Croci, Li, Cameron, Jaggi, Alistarh, Hoefler, and Hensman]{ashkboos2024quarot}
Ashkboos, S., Mohtashami, A., Croci, M.~L., Li, B., Cameron, P., Jaggi, M., Alistarh, D., Hoefler, T., and Hensman, J.
\newblock Quarot: Outlier-free 4-bit inference in rotated llms.
\newblock \emph{arXiv preprint arXiv:2404.00456}, 2024{\natexlab{b}}.

\bibitem[Bae et~al.(2024)Bae, Fisch, Harutyunyan, Ji, Kim, and Schuster]{bae2024relaxed}
Bae, S., Fisch, A., Harutyunyan, H., Ji, Z., Kim, S., and Schuster, T.
\newblock Relaxed recursive transformers: Effective parameter sharing with layer-wise lora.
\newblock \emph{arXiv preprint arXiv:2410.20672}, 2024.

\bibitem[Bai et~al.(2023)Bai, Bai, Chu, Cui, Dang, Deng, Fan, Ge, Han, Huang, et~al.]{bai2023qwen}
Bai, J., Bai, S., Chu, Y., Cui, Z., Dang, K., Deng, X., Fan, Y., Ge, W., Han, Y., Huang, F., et~al.
\newblock Qwen technical report.
\newblock \emph{arXiv preprint arXiv:2309.16609}, 2023.

\bibitem[Ben~Noach \& Goldberg(2020)Ben~Noach and Goldberg]{ben-noach-goldberg-2020-compressing}
Ben~Noach, M. and Goldberg, Y.
\newblock Compressing pre-trained language models by matrix decomposition.
\newblock In Wong, K.-F., Knight, K., and Wu, H. (eds.), \emph{Proceedings of the 1st Conference of the Asia-Pacific Chapter of the Association for Computational Linguistics and the 10th International Joint Conference on Natural Language Processing}, pp.\  884--889, Suzhou, China, December 2020. Association for Computational Linguistics.
\newblock \doi{10.18653/v1/2020.aacl-main.88}.
\newblock URL \url{https://aclanthology.org/2020.aacl-main.88/}.

\bibitem[Bian et~al.(2021)Bian, Huang, Cai, Yuan, and Church]{bian-etal-2021-attention}
Bian, Y., Huang, J., Cai, X., Yuan, J., and Church, K.
\newblock On attention redundancy: A comprehensive study.
\newblock In Toutanova, K., Rumshisky, A., Zettlemoyer, L., Hakkani-Tur, D., Beltagy, I., Bethard, S., Cotterell, R., Chakraborty, T., and Zhou, Y. (eds.), \emph{Proceedings of the 2021 Conference of the North American Chapter of the Association for Computational Linguistics: Human Language Technologies}, pp.\  930--945, Online, June 2021. Association for Computational Linguistics.
\newblock \doi{10.18653/v1/2021.naacl-main.72}.
\newblock URL \url{https://aclanthology.org/2021.naacl-main.72}.

\bibitem[Brown et~al.(2020)Brown, Mann, Ryder, Subbiah, Kaplan, Dhariwal, Neelakantan, Shyam, Sastry, Askell, Agarwal, Herbert{-}Voss, Krueger, Henighan, Child, Ramesh, Ziegler, Wu, Winter, Hesse, Chen, Sigler, Litwin, Gray, Chess, Clark, Berner, McCandlish, Radford, Sutskever, and Amodei]{BrownGPT3}
Brown, T.~B., Mann, B., Ryder, N., Subbiah, M., Kaplan, J., Dhariwal, P., Neelakantan, A., Shyam, P., Sastry, G., Askell, A., Agarwal, S., Herbert{-}Voss, A., Krueger, G., Henighan, T., Child, R., Ramesh, A., Ziegler, D.~M., Wu, J., Winter, C., Hesse, C., Chen, M., Sigler, E., Litwin, M., Gray, S., Chess, B., Clark, J., Berner, C., McCandlish, S., Radford, A., Sutskever, I., and Amodei, D.
\newblock Language models are few-shot learners.
\newblock In Larochelle, H., Ranzato, M., Hadsell, R., Balcan, M., and Lin, H. (eds.), \emph{Advances in Neural Information Processing Systems 33: Annual Conference on Neural Information Processing Systems 2020, NeurIPS 2020, December 6-12, 2020, virtual}, 2020.
\newblock URL \url{https://proceedings.neurips.cc/paper/2020/hash/1457c0d6bfcb4967418bfb8ac142f64a-Abstract.html}.

\bibitem[B{\"u}y{\"u}kaky{\"u}z(2024)]{buyukakyuz2024olora}
B{\"u}y{\"u}kaky{\"u}z, K.
\newblock Olora: Orthonormal low-rank adaptation of large language models.
\newblock \emph{arXiv preprint arXiv:2406.01775}, 2024.

\bibitem[Cai et~al.(2010)Cai, Cand{\`e}s, and Shen]{cai2010singular}
Cai, J.-F., Cand{\`e}s, E.~J., and Shen, Z.
\newblock A singular value thresholding algorithm for matrix completion.
\newblock \emph{SIAM Journal on optimization}, 20\penalty0 (4):\penalty0 1956--1982, 2010.

\bibitem[Cai et~al.(2024)Cai, Zhang, Gao, Liu, Liu, Lu, Xiong, Dong, Chang, Hu, and Xiao]{Cai2024PyramidKV}
Cai, Z., Zhang, Y., Gao, B., Liu, Y., Liu, T., Lu, K., Xiong, W., Dong, Y., Chang, B., Hu, J., and Xiao, W.
\newblock Pyramidkv: Dynamic {KV} cache compression based on pyramidal information funneling.
\newblock \emph{CoRR}, abs/2406.02069, 2024.
\newblock \doi{10.48550/ARXIV.2406.02069}.
\newblock URL \url{https://doi.org/10.48550/arXiv.2406.02069}.

\bibitem[Clark et~al.(2018)Clark, Cowhey, Etzioni, Khot, Sabharwal, Schoenick, and Tafjord]{clark2018think}
Clark, P., Cowhey, I., Etzioni, O., Khot, T., Sabharwal, A., Schoenick, C., and Tafjord, O.
\newblock Think you have solved question answering? try arc, the ai2 reasoning challenge.
\newblock \emph{arXiv preprint arXiv:1803.05457}, 2018.

\bibitem[Dehghani et~al.(2019)Dehghani, Gouws, Vinyals, Uszkoreit, and Kaiser]{dehghani2018universal}
Dehghani, M., Gouws, S., Vinyals, O., Uszkoreit, J., and Kaiser, L.
\newblock Universal transformers.
\newblock In \emph{International Conference on Learning Representations}, 2019.
\newblock URL \url{https://openreview.net/forum?id=HyzdRiR9Y7}.

\bibitem[Denton et~al.(2014)Denton, Zaremba, Bruna, LeCun, and Fergus]{NIPS2014_2afe4567}
Denton, E.~L., Zaremba, W., Bruna, J., LeCun, Y., and Fergus, R.
\newblock Exploiting linear structure within convolutional networks for efficient evaluation.
\newblock In Ghahramani, Z., Welling, M., Cortes, C., Lawrence, N., and Weinberger, K. (eds.), \emph{Advances in Neural Information Processing Systems}, volume~27. Curran Associates, Inc., 2014.
\newblock URL \url{https://proceedings.neurips.cc/paper_files/paper/2014/file/2afe4567e1bf64d32a5527244d104cea-Paper.pdf}.

\bibitem[Dettmers et~al.(2022)Dettmers, Lewis, Belkada, and Zettlemoyer]{dettmers2022gpt3}
Dettmers, T., Lewis, M., Belkada, Y., and Zettlemoyer, L.
\newblock Gpt3. int8 (): 8-bit matrix multiplication for transformers at scale.
\newblock \emph{Advances in Neural Information Processing Systems}, 35:\penalty0 30318--30332, 2022.

\bibitem[Ding et~al.(2023)Ding, Chen, Xu, Qin, Zheng, Hu, Liu, Sun, and Zhou]{ding2023enhancing}
Ding, N., Chen, Y., Xu, B., Qin, Y., Zheng, Z., Hu, S., Liu, Z., Sun, M., and Zhou, B.
\newblock Enhancing chat language models by scaling high-quality instructional conversations.
\newblock \emph{arXiv preprint arXiv:2305.14233}, 2023.

\bibitem[Dubey et~al.(2024)Dubey, Jauhri, Pandey, Kadian, Al-Dahle, Letman, Mathur, Schelten, Yang, Fan, et~al.]{dubey2024llama}
Dubey, A., Jauhri, A., Pandey, A., Kadian, A., Al-Dahle, A., Letman, A., Mathur, A., Schelten, A., Yang, A., Fan, A., et~al.
\newblock The llama 3 herd of models.
\newblock \emph{arXiv preprint arXiv:2407.21783}, 2024.

\bibitem[Eigen et~al.(2014)Eigen, Ranzato, and Sutskever]{EigenRS13MoE}
Eigen, D., Ranzato, M., and Sutskever, I.
\newblock Learning factored representations in a deep mixture of experts.
\newblock In Bengio, Y. and LeCun, Y. (eds.), \emph{2nd International Conference on Learning Representations, {ICLR} 2014, Banff, AB, Canada, April 14-16, 2014, Workshop Track Proceedings}, 2014.
\newblock URL \url{http://arxiv.org/abs/1312.4314}.

\bibitem[Fedus et~al.(2022)Fedus, Dean, and Zoph]{fedus2022reviewMoE}
Fedus, W., Dean, J., and Zoph, B.
\newblock A review of sparse expert models in deep learning.
\newblock \emph{arXiv preprint arXiv:2209.01667}, 2022.

\bibitem[Frankle \& Carbin(2019)Frankle and Carbin]{frankle2018the}
Frankle, J. and Carbin, M.
\newblock The lottery ticket hypothesis: Finding sparse, trainable neural networks.
\newblock In \emph{International Conference on Learning Representations}, 2019.
\newblock URL \url{https://openreview.net/forum?id=rJl-b3RcF7}.

\bibitem[Frankle et~al.(2020)Frankle, Dziugaite, Roy, and Carbin]{DBLP:conf/icml/FrankleD0C20}
Frankle, J., Dziugaite, G.~K., Roy, D.~M., and Carbin, M.
\newblock Linear mode connectivity and the lottery ticket hypothesis.
\newblock In \emph{Proceedings of the 37th International Conference on Machine Learning, {ICML} 2020, 13-18 July 2020, Virtual Event}, volume 119 of \emph{Proceedings of Machine Learning Research}, pp.\  3259--3269. {PMLR}, 2020.
\newblock URL \url{http://proceedings.mlr.press/v119/frankle20a.html}.

\bibitem[Frantar \& Alistarh(2022)Frantar and Alistarh]{FrantarA22}
Frantar, E. and Alistarh, D.
\newblock Optimal brain compression: {A} framework for accurate post-training quantization and pruning.
\newblock In Koyejo, S., Mohamed, S., Agarwal, A., Belgrave, D., Cho, K., and Oh, A. (eds.), \emph{Advances in Neural Information Processing Systems 35: Annual Conference on Neural Information Processing Systems 2022, NeurIPS 2022, New Orleans, LA, USA, November 28 - December 9, 2022}, 2022.
\newblock URL \url{http://papers.nips.cc/paper\_files/paper/2022/hash/1caf09c9f4e6b0150b06a07e77f2710c-Abstract-Conference.html}.

\bibitem[Frantar \& Alistarh(2023{\natexlab{a}})Frantar and Alistarh]{FrantarSparseGPT}
Frantar, E. and Alistarh, D.
\newblock Sparsegpt: Massive language models can be accurately pruned in one-shot.
\newblock In Krause, A., Brunskill, E., Cho, K., Engelhardt, B., Sabato, S., and Scarlett, J. (eds.), \emph{International Conference on Machine Learning, {ICML} 2023, 23-29 July 2023, Honolulu, Hawaii, {USA}}, volume 202 of \emph{Proceedings of Machine Learning Research}, pp.\  10323--10337. {PMLR}, 2023{\natexlab{a}}.
\newblock URL \url{https://proceedings.mlr.press/v202/frantar23a.html}.

\bibitem[Frantar \& Alistarh(2023{\natexlab{b}})Frantar and Alistarh]{frantar2023sparsegpt}
Frantar, E. and Alistarh, D.
\newblock Sparsegpt: Massive language models can be accurately pruned in one-shot.
\newblock In \emph{International Conference on Machine Learning}, pp.\  10323--10337. PMLR, 2023{\natexlab{b}}.

\bibitem[Frantar et~al.(2022)Frantar, Ashkboos, Hoefler, and Alistarh]{Frantar23gptq}
Frantar, E., Ashkboos, S., Hoefler, T., and Alistarh, D.
\newblock {GPTQ:} accurate post-training quantization for generative pre-trained transformers.
\newblock \emph{CoRR}, abs/2210.17323, 2022.
\newblock \doi{10.48550/ARXIV.2210.17323}.
\newblock URL \url{https://doi.org/10.48550/arXiv.2210.17323}.

\bibitem[Gale et~al.(2019)Gale, Elsen, and Hooker]{Trevor2019sparsity}
Gale, T., Elsen, E., and Hooker, S.
\newblock The state of sparsity in deep neural networks.
\newblock \emph{CoRR}, abs/1902.09574, 2019.
\newblock URL \url{http://arxiv.org/abs/1902.09574}.

\bibitem[Gao et~al.(2024)Gao, Tow, Abbasi, Biderman, Black, DiPofi, Foster, Golding, Hsu, Le~Noac'h, Li, McDonell, Muennighoff, Ociepa, Phang, Reynolds, Schoelkopf, Skowron, Sutawika, Tang, Thite, Wang, Wang, and Zou]{eval-harness}
Gao, L., Tow, J., Abbasi, B., Biderman, S., Black, S., DiPofi, A., Foster, C., Golding, L., Hsu, J., Le~Noac'h, A., Li, H., McDonell, K., Muennighoff, N., Ociepa, C., Phang, J., Reynolds, L., Schoelkopf, H., Skowron, A., Sutawika, L., Tang, E., Thite, A., Wang, B., Wang, K., and Zou, A.
\newblock A framework for few-shot language model evaluation, 07 2024.
\newblock URL \url{https://zenodo.org/records/12608602}.

\bibitem[Gu et~al.(2024)Gu, Dong, Wei, and Huang]{gu2024minillm}
Gu, Y., Dong, L., Wei, F., and Huang, M.
\newblock Mini{LLM}: Knowledge distillation of large language models.
\newblock In \emph{The Twelfth International Conference on Learning Representations}, 2024.
\newblock URL \url{https://openreview.net/forum?id=5h0qf7IBZZ}.

\bibitem[Han et~al.(2015{\natexlab{a}})Han, Mao, and Dally]{han2015deep}
Han, S., Mao, H., and Dally, W.~J.
\newblock Deep compression: Compressing deep neural networks with pruning, trained quantization and huffman coding.
\newblock \emph{arXiv preprint arXiv:1510.00149}, 2015{\natexlab{a}}.

\bibitem[Han et~al.(2015{\natexlab{b}})Han, Pool, Tran, and Dally]{Han2015}
Han, S., Pool, J., Tran, J., and Dally, W.~J.
\newblock Learning both weights and connections for efficient neural networks.
\newblock \emph{CoRR}, abs/1506.02626, 2015{\natexlab{b}}.
\newblock URL \url{http://arxiv.org/abs/1506.02626}.

\bibitem[He et~al.(2024)He, Sun, Shen, and Li]{he2024matterstransformersattentionneeded}
He, S., Sun, G., Shen, Z., and Li, A.
\newblock What matters in transformers? not all attention is needed.
\newblock \emph{CoRR}, abs/2406.15786, 2024.
\newblock URL \url{https://doi.org/10.48550/arXiv.2406.15786}.

\bibitem[Hinton(2015)]{hinton2015distilling}
Hinton, G.
\newblock Distilling the knowledge in a neural network.
\newblock \emph{arXiv preprint arXiv:1503.02531}, 2015.

\bibitem[Hoefler et~al.(2021)Hoefler, Alistarh, Ben{-}Nun, Dryden, and Peste]{HoeflerABDP21}
Hoefler, T., Alistarh, D., Ben{-}Nun, T., Dryden, N., and Peste, A.
\newblock Sparsity in deep learning: Pruning and growth for efficient inference and training in neural networks.
\newblock \emph{J. Mach. Learn. Res.}, 22:\penalty0 241:1--241:124, 2021.
\newblock URL \url{https://jmlr.org/papers/v22/21-0366.html}.

\bibitem[Hoffmann et~al.(2022)Hoffmann, Borgeaud, Mensch, Buchatskaya, Cai, Rutherford, de~Las~Casas, Hendricks, Welbl, Clark, Hennigan, Noland, Millican, van~den Driessche, Damoc, Guy, Osindero, Simonyan, Elsen, Rae, Vinyals, and Sifre]{Hoffmann22Chinchila}
Hoffmann, J., Borgeaud, S., Mensch, A., Buchatskaya, E., Cai, T., Rutherford, E., de~Las~Casas, D., Hendricks, L.~A., Welbl, J., Clark, A., Hennigan, T., Noland, E., Millican, K., van~den Driessche, G., Damoc, B., Guy, A., Osindero, S., Simonyan, K., Elsen, E., Rae, J.~W., Vinyals, O., and Sifre, L.
\newblock Training compute-optimal large language models.
\newblock \emph{CoRR}, abs/2203.15556, 2022.
\newblock \doi{10.48550/ARXIV.2203.15556}.
\newblock URL \url{https://doi.org/10.48550/arXiv.2203.15556}.

\bibitem[Hu et~al.(2021)Hu, Shen, Wallis, Allen-Zhu, Li, Wang, Wang, and Chen]{hu2021lora}
Hu, E.~J., Shen, Y., Wallis, P., Allen-Zhu, Z., Li, Y., Wang, S., Wang, L., and Chen, W.
\newblock Lora: Low-rank adaptation of large language models.
\newblock \emph{arXiv preprint arXiv:2106.09685}, 2021.

\bibitem[Hu et~al.(2022)Hu, yelong shen, Wallis, Allen-Zhu, Li, Wang, Wang, and Chen]{hu2022lora}
Hu, E.~J., yelong shen, Wallis, P., Allen-Zhu, Z., Li, Y., Wang, S., Wang, L., and Chen, W.
\newblock Lo{RA}: Low-rank adaptation of large language models.
\newblock In \emph{International Conference on Learning Representations}, 2022.
\newblock URL \url{https://openreview.net/forum?id=nZeVKeeFYf9}.

\bibitem[Hyeon-Woo et~al.(2021)Hyeon-Woo, Ye-Bin, and Oh]{hyeon2021fedpara}
Hyeon-Woo, N., Ye-Bin, M., and Oh, T.-H.
\newblock Fedpara: Low-rank hadamard product for communication-efficient federated learning.
\newblock \emph{arXiv preprint arXiv:2108.06098}, 2021.

\bibitem[Jacobs et~al.(1991)Jacobs, Jordan, Nowlan, and Hinton]{JacobsJNH91MOE}
Jacobs, R.~A., Jordan, M.~I., Nowlan, S.~J., and Hinton, G.~E.
\newblock Adaptive mixtures of local experts.
\newblock \emph{Neural Comput.}, 3\penalty0 (1):\penalty0 79--87, 1991.
\newblock \doi{10.1162/NECO.1991.3.1.79}.
\newblock URL \url{https://doi.org/10.1162/neco.1991.3.1.79}.

\bibitem[Jiang et~al.(2024)Jiang, Sablayrolles, Roux, Mensch, Savary, Bamford, Chaplot, Casas, Hanna, Bressand, et~al.]{jiang2024mixtral}
Jiang, A.~Q., Sablayrolles, A., Roux, A., Mensch, A., Savary, B., Bamford, C., Chaplot, D.~S., Casas, D. d.~l., Hanna, E.~B., Bressand, F., et~al.
\newblock Mixtral of experts.
\newblock \emph{arXiv preprint arXiv:2401.04088}, 2024.

\bibitem[Kaplan et~al.(2020)Kaplan, McCandlish, Henighan, Brown, Chess, Child, Gray, Radford, Wu, and Amodei]{Kaplan20Scalinglaw}
Kaplan, J., McCandlish, S., Henighan, T., Brown, T.~B., Chess, B., Child, R., Gray, S., Radford, A., Wu, J., and Amodei, D.
\newblock Scaling laws for neural language models.
\newblock \emph{CoRR}, abs/2001.08361, 2020.
\newblock URL \url{https://arxiv.org/abs/2001.08361}.

\bibitem[Lan et~al.(2020)Lan, Chen, Goodman, Gimpel, Sharma, and Soricut]{Lan2020ALBERT}
Lan, Z., Chen, M., Goodman, S., Gimpel, K., Sharma, P., and Soricut, R.
\newblock Albert: A lite bert for self-supervised learning of language representations.
\newblock In \emph{International Conference on Learning Representations}, 2020.
\newblock URL \url{https://openreview.net/forum?id=H1eA7AEtvS}.

\bibitem[Leviathan et~al.(2023)Leviathan, Kalman, and Matias]{LeviathanKM23specDecoding}
Leviathan, Y., Kalman, M., and Matias, Y.
\newblock Fast inference from transformers via speculative decoding.
\newblock In Krause, A., Brunskill, E., Cho, K., Engelhardt, B., Sabato, S., and Scarlett, J. (eds.), \emph{International Conference on Machine Learning, {ICML} 2023, 23-29 July 2023, Honolulu, Hawaii, {USA}}, volume 202 of \emph{Proceedings of Machine Learning Research}, pp.\  19274--19286. {PMLR}, 2023.
\newblock URL \url{https://proceedings.mlr.press/v202/leviathan23a.html}.

\bibitem[Li et~al.(2016)Li, Liang, and Risteski]{li2016recovery}
Li, Y., Liang, Y., and Risteski, A.
\newblock Recovery guarantee of weighted low-rank approximation via alternating minimization.
\newblock In \emph{International Conference on Machine Learning}, pp.\  2358--2367. PMLR, 2016.

\bibitem[Li et~al.(2018)Li, Ma, and Zhang]{li2018algorithmic}
Li, Y., Ma, T., and Zhang, H.
\newblock Algorithmic regularization in over-parameterized matrix sensing and neural networks with quadratic activations.
\newblock In \emph{Conference On Learning Theory}, pp.\  2--47. PMLR, 2018.

\bibitem[Li et~al.(2024)Li, Wei, Zhang, and Zhang]{LiW0024eagle2}
Li, Y., Wei, F., Zhang, C., and Zhang, H.
\newblock {EAGLE-2:} faster inference of language models with dynamic draft trees.
\newblock In Al{-}Onaizan, Y., Bansal, M., and Chen, Y. (eds.), \emph{Proceedings of the 2024 Conference on Empirical Methods in Natural Language Processing, {EMNLP} 2024, Miami, FL, USA, November 12-16, 2024}, pp.\  7421--7432. Association for Computational Linguistics, 2024.
\newblock URL \url{https://aclanthology.org/2024.emnlp-main.422}.

\bibitem[Liu et~al.(2023)Liu, Desai, Liao, Wang, Xie, Xu, Kyrillidis, and Shrivastava]{LiuDL2023Scissorhands}
Liu, Z., Desai, A., Liao, F., Wang, W., Xie, V., Xu, Z., Kyrillidis, A., and Shrivastava, A.
\newblock Scissorhands: Exploiting the persistence of importance hypothesis for {LLM} {KV} cache compression at test time.
\newblock In Oh, A., Naumann, T., Globerson, A., Saenko, K., Hardt, M., and Levine, S. (eds.), \emph{Advances in Neural Information Processing Systems 36: Annual Conference on Neural Information Processing Systems 2023, NeurIPS 2023, New Orleans, LA, USA, December 10 - 16, 2023}, 2023.
\newblock URL \url{http://papers.nips.cc/paper\_files/paper/2023/hash/a452a7c6c463e4ae8fbdc614c6e983e6-Abstract-Conference.html}.

\bibitem[Liu et~al.(2024)Liu, Zhao, Iandola, Lai, Tian, Fedorov, Xiong, Chang, Shi, Krishnamoorthi, et~al.]{MobileLLM}
Liu, Z., Zhao, C., Iandola, F., Lai, C., Tian, Y., Fedorov, I., Xiong, Y., Chang, E., Shi, Y., Krishnamoorthi, R., et~al.
\newblock Mobilellm: Optimizing sub-billion parameter language models for on-device use cases.
\newblock In \emph{International Conference on Machine Learning}, 2024.

\bibitem[Ma et~al.(2023)Ma, Fang, and Wang]{ma2023llmpruner}
Ma, X., Fang, G., and Wang, X.
\newblock Llm-pruner: On the structural pruning of large language models.
\newblock \emph{Advances in neural information processing systems}, 36:\penalty0 21702--21720, 2023.

\bibitem[Mahabadi et~al.(2021)Mahabadi, Henderson, and Ruder]{Mahabadi21compacter}
Mahabadi, R.~K., Henderson, J., and Ruder, S.
\newblock Compacter: Efficient low-rank hypercomplex adapter layers.
\newblock In Ranzato, M., Beygelzimer, A., Dauphin, Y.~N., Liang, P., and Vaughan, J.~W. (eds.), \emph{Advances in Neural Information Processing Systems 34: Annual Conference on Neural Information Processing Systems 2021, NeurIPS 2021, December 6-14, 2021, virtual}, pp.\  1022--1035, 2021.
\newblock URL \url{https://proceedings.neurips.cc/paper/2021/hash/081be9fdff07f3bc808f935906ef70c0-Abstract.html}.

\bibitem[Mangrulkar et~al.(2022)Mangrulkar, Gugger, Debut, Belkada, Paul, and Bossan]{peft}
Mangrulkar, S., Gugger, S., Debut, L., Belkada, Y., Paul, S., and Bossan, B.
\newblock Peft: State-of-the-art parameter-efficient fine-tuning methods.
\newblock \url{https://github.com/huggingface/peft}, 2022.

\bibitem[Men et~al.(2024)Men, Xu, Zhang, Wang, Lin, Lu, Han, and Chen]{men2024shortgpt}
Men, X., Xu, M., Zhang, Q., Wang, B., Lin, H., Lu, Y., Han, X., and Chen, W.
\newblock Shortgpt: Layers in large language models are more redundant than you expect.
\newblock \emph{arXiv preprint arXiv:2403.03853}, 2024.

\bibitem[Meng et~al.(2024)Meng, Wang, and Zhang]{meng2024pissa}
Meng, F., Wang, Z., and Zhang, M.
\newblock Pissa: Principal singular values and singular vectors adaptation of large language models.
\newblock \emph{arXiv preprint arXiv:2404.02948}, 2024.

\bibitem[Michael H.~Zhu(2018)]{h.2018to}
Michael H.~Zhu, S.~G.
\newblock To prune, or not to prune: Exploring the efficacy of pruning for model compression, 2018.
\newblock URL \url{https://openreview.net/forum?id=S1lN69AT-}.

\bibitem[Mishra et~al.(2021)Mishra, Latorre, Pool, Stosic, Stosic, Venkatesh, Yu, and Micikevicius]{Mishra2021}
Mishra, A.~K., Latorre, J.~A., Pool, J., Stosic, D., Stosic, D., Venkatesh, G., Yu, C., and Micikevicius, P.
\newblock Accelerating sparse deep neural networks.
\newblock \emph{CoRR}, abs/2104.08378, 2021.
\newblock URL \url{https://arxiv.org/abs/2104.08378}.

\bibitem[Muralidharan et~al.(2024)Muralidharan, Sreenivas, Joshi, Chochowski, Patwary, Shoeybi, Catanzaro, Kautz, and Molchanov]{muralidharan2024compact}
Muralidharan, S., Sreenivas, S.~T., Joshi, R.~B., Chochowski, M., Patwary, M., Shoeybi, M., Catanzaro, B., Kautz, J., and Molchanov, P.
\newblock Compact language models via pruning and knowledge distillation.
\newblock In \emph{The Thirty-eighth Annual Conference on Neural Information Processing Systems}, 2024.
\newblock URL \url{https://openreview.net/forum?id=9U0nLnNMJ7}.

\bibitem[Oymak et~al.(2019)Oymak, Fabian, Li, and Soltanolkotabi]{oymak2019generalization}
Oymak, S., Fabian, Z., Li, M., and Soltanolkotabi, M.
\newblock Generalization guarantees for neural networks via harnessing the low-rank structure of the jacobian.
\newblock \emph{arXiv preprint arXiv:1906.05392}, 2019.

\bibitem[Paischer et~al.(2024)Paischer, Hauzenberger, Schmied, Alkin, Deisenroth, and Hochreiter]{paischer2024one}
Paischer, F., Hauzenberger, L., Schmied, T., Alkin, B., Deisenroth, M.~P., and Hochreiter, S.
\newblock One initialization to rule them all: Fine-tuning via explained variance adaptation.
\newblock \emph{arXiv preprint arXiv:2410.07170}, 2024.

\bibitem[Pan et~al.(2024)Pan, Wu, Jiang, Xia, Luo, Zhang, Lin, R{\"u}hle, Yang, Lin, Zhao, Qiu, and Zhang]{pan-etal-2024-llmlingua}
Pan, Z., Wu, Q., Jiang, H., Xia, M., Luo, X., Zhang, J., Lin, Q., R{\"u}hle, V., Yang, Y., Lin, C.-Y., Zhao, H.~V., Qiu, L., and Zhang, D.
\newblock {LLML}ingua-2: Data distillation for efficient and faithful task-agnostic prompt compression.
\newblock In Ku, L.-W., Martins, A., and Srikumar, V. (eds.), \emph{Findings of the Association for Computational Linguistics: ACL 2024}, pp.\  963--981, Bangkok, Thailand, August 2024. Association for Computational Linguistics.
\newblock \doi{10.18653/v1/2024.findings-acl.57}.
\newblock URL \url{https://aclanthology.org/2024.findings-acl.57/}.

\bibitem[Rohan~Taori \& Hashimoto(2013)Rohan~Taori and Hashimoto]{alpaca}
Rohan~Taori, Ishaan~Gulrajani, T. Z. Y. D. X. L. C. G. P.~L. and Hashimoto, T.~B.
\newblock Stanford alpaca: An instruction-following llama model, 2013.
\newblock URL \url{https://github.com/tatsu-lab/stanford_alpaca}.

\bibitem[Sakaguchi et~al.(2021)Sakaguchi, Bras, Bhagavatula, and Choi]{sakaguchi2021winogrande}
Sakaguchi, K., Bras, R.~L., Bhagavatula, C., and Choi, Y.
\newblock Winogrande: An adversarial winograd schema challenge at scale.
\newblock \emph{Communications of the ACM}, 64\penalty0 (9):\penalty0 99--106, 2021.

\bibitem[Salinas et~al.(2022)Salinas, Seeger, Klein, Perrone, Wistuba, and Archambeau]{salinas2022syne}
Salinas, D., Seeger, M., Klein, A., Perrone, V., Wistuba, M., and Archambeau, C.
\newblock {Syne Tune}: A library for large scale hyperparameter tuning and reproducible research.
\newblock In \emph{International Conference on Automated Machine Learning, AutoML 2022}, 2022.
\newblock URL \url{https://proceedings.mlr.press/v188/salinas22a.html}.

\bibitem[Sanh et~al.(2020)Sanh, Wolf, and Rush]{Sanh0R20}
Sanh, V., Wolf, T., and Rush, A.~M.
\newblock Movement pruning: Adaptive sparsity by fine-tuning.
\newblock In Larochelle, H., Ranzato, M., Hadsell, R., Balcan, M., and Lin, H. (eds.), \emph{Advances in Neural Information Processing Systems 33: Annual Conference on Neural Information Processing Systems 2020, NeurIPS 2020, December 6-12, 2020, virtual}, 2020.
\newblock URL \url{https://proceedings.neurips.cc/paper/2020/hash/eae15aabaa768ae4a5993a8a4f4fa6e4-Abstract.html}.

\bibitem[Shazeer et~al.(2017)Shazeer, Mirhoseini, Maziarz, Davis, Le, Hinton, and Dean]{ShazeerMMDLHD17MoE}
Shazeer, N., Mirhoseini, A., Maziarz, K., Davis, A., Le, Q.~V., Hinton, G.~E., and Dean, J.
\newblock Outrageously large neural networks: The sparsely-gated mixture-of-experts layer.
\newblock In \emph{5th International Conference on Learning Representations, {ICLR} 2017, Toulon, France, April 24-26, 2017, Conference Track Proceedings}. OpenReview.net, 2017.
\newblock URL \url{https://openreview.net/forum?id=B1ckMDqlg}.

\bibitem[Shoaib Ahmed~Siddiqui(2024)]{deeperLook}
Shoaib Ahmed~Siddiqui, Xin~Dong, G. H. T. B. J. K. D. K. P.~M.
\newblock A deeper look at depth pruning of llms.
\newblock \emph{arXiv preprint arXiv:2407.16286}, 2024.

\bibitem[Singh \& Alistarh(2020)Singh and Alistarh]{SinghAlistarh20}
Singh, S.~P. and Alistarh, D.
\newblock Woodfisher: Efficient second-order approximation for neural network compression.
\newblock In Larochelle, H., Ranzato, M., Hadsell, R., Balcan, M., and Lin, H. (eds.), \emph{Advances in Neural Information Processing Systems 33: Annual Conference on Neural Information Processing Systems 2020, NeurIPS 2020, December 6-12, 2020, virtual}, 2020.
\newblock URL \url{https://proceedings.neurips.cc/paper/2020/hash/d1ff1ec86b62cd5f3903ff19c3a326b2-Abstract.html}.

\bibitem[Sun et~al.(2024)Sun, Liu, Bair, and Kolter]{Sun24pruning}
Sun, M., Liu, Z., Bair, A., and Kolter, J.~Z.
\newblock A simple and effective pruning approach for large language models.
\newblock In \emph{The Twelfth International Conference on Learning Representations, {ICLR} 2024, Vienna, Austria, May 7-11, 2024}. OpenReview.net, 2024.
\newblock URL \url{https://openreview.net/forum?id=PxoFut3dWW}.

\bibitem[Team et~al.(2024)Team, Mesnard, Hardin, Dadashi, Bhupatiraju, Pathak, Sifre, Rivi{\`e}re, Kale, Love, et~al.]{team2024gemma}
Team, G., Mesnard, T., Hardin, C., Dadashi, R., Bhupatiraju, S., Pathak, S., Sifre, L., Rivi{\`e}re, M., Kale, M.~S., Love, J., et~al.
\newblock Gemma: Open models based on gemini research and technology.
\newblock \emph{arXiv preprint arXiv:2403.08295}, 2024.

\bibitem[Tukan et~al.(2020)Tukan, Maalouf, Weksler, and Feldman]{tukan2020compressed}
Tukan, M., Maalouf, A., Weksler, M., and Feldman, D.
\newblock Compressed deep networks: Goodbye svd, hello robust low-rank approximation.
\newblock \emph{arXiv preprint arXiv:2009.05647}, 2020.

\bibitem[Vaswani(2017)]{vaswani2017attention}
Vaswani, A.
\newblock Attention is all you need.
\newblock \emph{Advances in Neural Information Processing Systems}, 2017.

\bibitem[Wang \& Li(2024)Wang and Li]{wang2024residualtransformer}
Wang, Y. and Li, J.
\newblock Residualtransformer: Residual low-rank learning with weight-sharing for transformer layers.
\newblock In \emph{ICASSP 2024-2024 IEEE International Conference on Acoustics, Speech and Signal Processing (ICASSP)}, pp.\  11161--11165. IEEE, 2024.

\bibitem[Wang et~al.(2024)Wang, Ma, Zhang, Ni, Chandra, Guo, Ren, Arulraj, He, Jiang, et~al.]{wang2024mmlu}
Wang, Y., Ma, X., Zhang, G., Ni, Y., Chandra, A., Guo, S., Ren, W., Arulraj, A., He, X., Jiang, Z., et~al.
\newblock Mmlu-pro: A more robust and challenging multi-task language understanding benchmark.
\newblock \emph{arXiv preprint arXiv:2406.01574}, 2024.

\bibitem[Xu et~al.(2020)Xu, Zhou, Ge, Wei, and Zhou]{xu2020berttheseus}
Xu, C., Zhou, W., Ge, T., Wei, F., and Zhou, M.
\newblock Bert-of-theseus: Compressing bert by progressive module replacing.
\newblock \emph{arXiv preprint arXiv:2002.02925}, 2020.

\bibitem[Xu et~al.(2024)Xu, Han, Yang, Wang, Zhu, Liu, Liu, and Che]{xu2024onebit}
Xu, Y., Han, X., Yang, Z., Wang, S., Zhu, Q., Liu, Z., Liu, W., and Che, W.
\newblock Onebit: Towards extremely low-bit large language models.
\newblock \emph{arXiv preprint arXiv:2402.11295}, 2024.

\bibitem[Yang et~al.(2024)Yang, Cao, and Zhao]{YangC024LaCo}
Yang, Y., Cao, Z., and Zhao, H.
\newblock Laco: Large language model pruning via layer collapse.
\newblock In Al{-}Onaizan, Y., Bansal, M., and Chen, Y. (eds.), \emph{Findings of the Association for Computational Linguistics: {EMNLP} 2024, Miami, Florida, USA, November 12-16, 2024}, pp.\  6401--6417. Association for Computational Linguistics, 2024.
\newblock URL \url{https://aclanthology.org/2024.findings-emnlp.372}.

\bibitem[YEH et~al.(2024)YEH, Hsieh, Gao, Yang, Oh, and Gong]{yeh2024navigating}
YEH, S.-Y., Hsieh, Y.-G., Gao, Z., Yang, B. B.~W., Oh, G., and Gong, Y.
\newblock Navigating text-to-image customization: From ly{CORIS} fine-tuning to model evaluation.
\newblock In \emph{The Twelfth International Conference on Learning Representations}, 2024.
\newblock URL \url{https://openreview.net/forum?id=wfzXa8e783}.

\bibitem[Yu et~al.(2020)Yu, Edunov, Tian, and Morcos]{Yu2020Playing}
Yu, H., Edunov, S., Tian, Y., and Morcos, A.~S.
\newblock Playing the lottery with rewards and multiple languages: lottery tickets in rl and nlp.
\newblock In \emph{International Conference on Learning Representations}, 2020.
\newblock URL \url{https://openreview.net/forum?id=S1xnXRVFwH}.

\bibitem[Zellers et~al.(2019)Zellers, Holtzman, Bisk, Farhadi, and Choi]{zellers-etal-2019-hellaswag}
Zellers, R., Holtzman, A., Bisk, Y., Farhadi, A., and Choi, Y.
\newblock {H}ella{S}wag: Can a machine really finish your sentence?
\newblock In Korhonen, A., Traum, D., and M{\`a}rquez, L. (eds.), \emph{Proceedings of the 57th Annual Meeting of the Association for Computational Linguistics}, pp.\  4791--4800, Florence, Italy, July 2019. Association for Computational Linguistics.
\newblock \doi{10.18653/v1/P19-1472}.
\newblock URL \url{https://aclanthology.org/P19-1472/}.

\bibitem[Zhang et~al.(2023)Zhang, Sheng, Zhou, Chen, Zheng, Cai, Song, Tian, R{\'{e}}, Barrett, Wang, and Chen]{Zhang2023H2O}
Zhang, Z., Sheng, Y., Zhou, T., Chen, T., Zheng, L., Cai, R., Song, Z., Tian, Y., R{\'{e}}, C., Barrett, C.~W., Wang, Z., and Chen, B.
\newblock {H2O:} heavy-hitter oracle for efficient generative inference of large language models.
\newblock In Oh, A., Naumann, T., Globerson, A., Saenko, K., Hardt, M., and Levine, S. (eds.), \emph{Advances in Neural Information Processing Systems 36: Annual Conference on Neural Information Processing Systems 2023, NeurIPS 2023, New Orleans, LA, USA, December 10 - 16, 2023}, 2023.
\newblock URL \url{http://papers.nips.cc/paper\_files/paper/2023/hash/6ceefa7b15572587b78ecfcebb2827f8-Abstract-Conference.html}.

\end{thebibliography}
\bibliographystyle{icml2025}

\newpage
\appendix
\onecolumn

\begin{table*}[t]
\caption{Comparison of Delta-Models to SLMs of similar sizes (perplexity)}
\label{Stage1}
\vskip 0.15in
\begin{center}
\begin{small}
\begin{tabular}{lcccccr}
\toprule
Model & \# Parameters & Compression \% & Dataset &  PPL  \\
\midrule
\textbf{Phi 3.5} & 3.82B & & Alpaca & \textbf{2.96} \\
Llama 3.2 & 3.2B & & Alpaca & 4.25 \\
Qwen 2.5 & 3.2B & & Alpaca &  3.69 \\
\midrule
\textbf{\phicompressedname} (5 seq MLP $\delta$, r1000) & 3.35B & 12\% & Alpaca & \textbf{3.34} \\
\phicompressedname (6 seq MLP $\delta$, r1000) & 3.3B &13\% & Alpaca & 3.80 \\
\phicompressedname  (8 seq MLP $\delta$, r100) & 2.8B &26\% & Alpaca & 3.94 \\
\llamacompressedname (6 seq  MLP $\delta$, r100) & 2.62B &18\% & Alpaca & 5.35 \\
\llamacompressedname (7 seq MLP $\delta$, r100) & 2.52B &21\% & Alpaca & 5.43 \\
\llamacompressedname (9 seq MLP $\delta$, r1000) & 2.41B & 25\% & Alpaca & 6.00 \\
\midrule
Phi 3.5 & 3.82B & & Ultrachat & 8.22 \\
Llama 3 & 3.2B & & Ultrachat & 14.81 \\
\textbf{\phicompressedname} (5 seq MLP $\delta$, r1000) & 3.35B & 12\% & Ultrachat & \textbf{7.49} \\
\phicompressedname (8 seq MLP $\delta$, r100) & 2.9B & 24\% & Ultrachat & 9.19 \\
\llamacompressedname (7 seq MLP $\delta$, r100) & 2.51B & 19\% & Ultrachat & 18.32 \\
\llamacompressedname (9 seq MLP $\delta$, r1000) & 2.41B & 25\% & Ultrachat & 17.74 \\
\bottomrule
\end{tabular}
\end{small}
\end{center}
\vskip -0.1in
\end{table*}



\end{document}